\documentclass[runningheads]{llncs}

\usepackage[T1]{fontenc}
\usepackage{graphicx}
\usepackage{amsmath}
\usepackage{amssymb}
\usepackage[nolist]{acronym}
\begin{acronym} 
 \acro{GAN}{Generative Adversarial Network}
 \acrodefplural{GAN}{Generative Adversarial Networks}
 \acro{DCGAN}{Deep Convolutional Generative Adversarial Network}
 \acrodefplural{DCGAN}{Deep Convolutional Generative Adversarial Networks}
 \acro{SNGAN}{Spectral Normalization Generative Adversarial Network}
 \acrodefplural{SNGAN}{Spectral Normalization Generative Adversarial Networks}
 \acro{SA-GAN}{Self-Attention Generative Adversarial Network}
 \acrodefplural{SA-GAN}{Self-Attention Generative Adversarial Networks}
 \acro{BigGAN}{Big Generative Adversarial Network}
 \acrodefplural{BigGAN}{Big Generative Adversarial Networks}
 \acro{WGAN}{Wasserstein Generative Adversarial Network}
 \acrodefplural{WGAN}{Wasserstein Generative Adversarial Networks}
 \acro{WGAN-GP}{Wasserstein Generative Adversarial Network with Gradient Penalty}
 \acro{DGMR}{Deep Generative Model of Rainfall}
 \acro{VAE}{Variational Autoencoder}
 \acrodefplural{VAE}{Variational Autoencoders}
 \acro{ASL}{Arterial Spin Labeling}
 \acro{MRI}{Magnetic Resonance Imaging}
 \acro{PET}{Positron Emission Tomography}
 \acro{ADNI}{Alzheimer’s Disease Neuroimaging Initiative}
 \acro{MSL}{Multiple Scale and Location}
 \acro{SVC}{Singular Value Clipping}
 \acro{LIDC}{The Lung Image Database Consortium image collection}
 \acro{TCIA}{The Cancer Imaging Archive}
 \acro{BRATS}{The Multimodal Brain Tumor Image Segmentation Benchmark}
 \acro{CT}{Computerized Tomography}
 \acro{HU}{Hounsfield Units}
 \acro{DICOM}{Digital Imaging and Communications in Medicine}
 \acro{FID}{Fréchet Inception Distance}
 \acro{EM}{Earth Mover}
 \acro{TWR}{Tournament Win Rate}
 \acro{MS-SSIM}{Multiple Scale Structural Similarity Index}
 \acro{bMMD$^2$}{batch-wise Maximum Mean Discrepancy$^2$}
 \acro{CAD}{Computer-Aided Diagnostics}
 \acro{TTUR}{Two Time-scale Update Rule}
\end{acronym}
\usepackage{hyperref}
\usepackage{color}

\usepackage{cite}

\begin{document}

\title{Explicit Temporal Embedding in Deep Generative Latent Models for Longitudinal Medical Image Synthesis}
\titlerunning{Temporal Embedding in Generative Latent Models}
\author{Julian Schön\inst{1,2} \and
Raghavendra Selvan\inst{1,3} \and
Lotte Nygård\inst{2} \and
Ivan Richter Vogelius\inst{2,4}\and
Jens Petersen\inst{1,2}}
\authorrunning{J. Schön et al.}
\institute{Department of Computer Science, University of Copenhagen, Denmark\\
\email{julian.e.s@di.ku.dk}
\and Department of Oncology, Rigshospitalet, Denmark\and
Department of Neuroscience, University of Copenhagen, Denmark\and
Faculty of Health and Medical Sciences, University of Copenhagen, Copenhagen, Denmark}
\maketitle              % typeset the header of the contribution
\begin{abstract}
Medical imaging plays a vital role in modern diagnostics and treatment. The temporal nature of disease or treatment progression often results in longitudinal data. Due to the cost and potential harm, acquiring large medical datasets necessary for deep learning can be difficult. Medical image synthesis could help mitigate this problem. However, until now, the availability of GANs capable of synthesizing longitudinal volumetric data has been limited. To address this, we use the recent advances in latent space-based image editing to propose a novel joint learning scheme to explicitly embed temporal dependencies in the latent space of GANs. This, in contrast to previous methods, allows us to synthesize continuous, smooth, and high-quality longitudinal volumetric data with limited supervision. We show the effectiveness of our approach on three datasets containing different longitudinal dependencies. Namely, modeling a simple image transformation, breathing motion, and tumor regression, all while showing minimal disentanglement. The implementation is made available online\footnote{\url{https://github.com/julschoen/Temp-GAN}}.
\end{abstract}

\section{Introduction}
The use of deep learning in the medical domain has increased recently but is limited by the need for large and well-labeled datasets \cite{Litjens17}. A potential mitigation to this problem is the use of synthetic data obtained using generative models such as \acp{GAN} \cite{GoodFellow14}, which has been shown to enhance medical deep learning algorithms \cite{Sandfort19}. Due to the natural temporal development of, e.g., disease progression or treatment monitoring, temporal data is gathered frequently in the medical domain. Prominent cases are neurodegenerative diseases such as Alzheimer's or cancer-related longitudinal data collected during radiotherapy. Combining longitudinal medical data and deep learning can allow for earlier and more accurate prognosis \cite{Wen20}, as well as disease modeling, such as tumor progression or regression \cite{HaiBin21}. \acp{GAN} have successfully been used to generate temporal data. They have shown promising results in video generation \cite{Saito17, Wang20}, precipitation forecasting \cite{Ravuri21}, and also medical temporal data generation \cite{Fan22, Abdullah22}. However, all previous approaches have been either on 2D data \cite{Saito17, Wang20, Ravuri21, Abdullah22} or have considered image-to-sequence or sequence-to-sequence generation tasks \cite{Wang20, Ravuri21, Fan22}. While temporal data generation can be done with image-to-sequence or sequence-to-sequence models, they do not allow for the generation of new sequences but rely on input data on which the generated data expands.\\
In recent years, a line of work has focused on the interpretability of \acp{GAN} by investigating them utilizing linear directions in their latent spaces that result in meaningful and interpretable image transformations \cite{Goetschalckx19, Jahanian20, Voynov20}. These works show that simple shifts of latent codes along a linear direction can result in powerful image transformations such as increasing memorability \cite{Goetschalckx19}, rotating the image subject \cite{Jahanian20}, or even background removal \cite{Voynov20}. However, these approaches operate on pre-trained \acp{GAN} and can only discover what is already captured by the learned representation.\\
The following summarises the main contributions of our work:
\begin{itemize}
    \item We propose a novel model, jointly training a \ac{GAN} and a direction in the latent space corresponding to any desired image transformation for which ordered data is available. To the best of our knowledge, our approach is the first to explicitly embed image transformations in the form of linear directions into the \ac{GAN} latent space during training. Furthermore, the proposed joint training procedure is model agnostic and works with any latent variable-based \ac{GAN}.
    \item We use the proposed framework to embed a linear direction corresponding to temporal changes in volumetric medical data. This allows the generation of longitudinal volumetric data for the first time without requiring input to the generator in the form of images or sequences of images. Furthermore, as the temporal sequence generation is based on a simple shift in the latent space, we can generate smooth and continuous sequences processing each time point individually, thereby lessening memory requirements compared to the processing of full sequences.
\end{itemize}

\section{Related Work}
\label{sec:related_work}
Our work considers concepts from different lines of prior research in generative modeling. In the following, we summarise this.

\subsubsection{Volumetric Data Synthesis} Despite the advances in natural image synthesis, there is a lack of usage and implementations of general-purpose state-of-the-art volumetric \acp{GAN} architectures. Existing volumetric \acp{GAN} either do not utilize current advances in \ac{GAN} architectures \cite{Wu16}, are task-specific \cite{Li21}, trade-off image resolution to allow for state-of-the-art model architectures \cite{Hong21} or are focusing on image-to-image generation tasks \cite{Lin20, Lan21}. More advanced architectures, such as \ac{SA-GAN} \cite{Zhang19}, which can be made more memory efficient by using residual blocks with bottlenecks as suggested with \ac{BigGAN} \cite{Brock19}, are well suited to volumetric data synthesis. However, while their use is common in the natural image domain, they are generally not used for volumetric data.

\subsubsection{Latent Based Image Editing} Latent-based image edits have been possible in \ac{GAN} latent spaces since the introduction of \ac{DCGAN} \cite{Radford16}. Currently, most approaches use linear directions, i.e., latent walks, in the latent space of pre-trained generators corresponding to interpretable image edits \cite{Goetschalckx19, Voynov20, Jahanian20}. The learned representation, however, does not necessarily contain any potentially desired image transformation. InfoGAN \cite{Chen16} mitigates this by jointly training the \ac{GAN} and an additional latent vector input to disentangle the learned representations. However, a potentially desired image transformation can not be explicitly enforced. In contrast, thanks to jointly training the generator and the desired embedding, we ensure that the desired edit is encoded in the latent space.

\subsubsection{Temporal Synthesis} Prior works on temporal synthesis have focused on video generation with \acp{GAN} \cite{Saito17, Wang20, Ravuri21}. This work was inspired by the shown viability of temporal latent embedding \cite{Saito17} and the use of two discriminators \cite{Saito17, Wang20, Ravuri21}. However, most approaches use generators conditioned on input images \cite{Wang20, Ravuri21}, limiting their ability to generate new sequences, the exception being Saito et al. \cite{Saito17}, which shows the viability of temporal generation based on latent models, they do so on natural images and expand the latent space for the temporal modeling. In contrast, our approach is the first to operate on volumetric data, and we make this possible by embedding the temporal component into the latent space directly rather than expanding it.\\
More similar approaches to ours have been introduced with TR-\ac{GAN} \cite{Fan22} and the work by Abdullah et al. \cite{Abdullah22}. TR-\ac{GAN} explores many-to-many predictions of \ac{MRI} using a single generator. Like the previous methods, TR-\ac{GAN} utilizes a generator conditioned on input sequences to predict temporal sequences. Thus, future time steps are directly generated from input. In contrast, our approach embeds temporal dependencies in the latent space. Thus, TR-\ac{GAN} relies on a sequential generator and cannot generate new sequences. Finally, Abdullah et al. \cite{Abdullah22} propose subdividing the latent space to embed temporal dependencies of medical images. However, they rely on 2D data, crops around the region of interest, and a-priori information on the time-dependent variable (e.g., accelerometer data). In contrast, our proposed method only requires the natural ordering of the temporal sequence, operates on entire volumes, and we choose linear latent embeddings.

\section{Methods}
\label{sec:methods}
\begin{figure}[h!]
    \centering
    \includegraphics[width=0.8\textwidth]{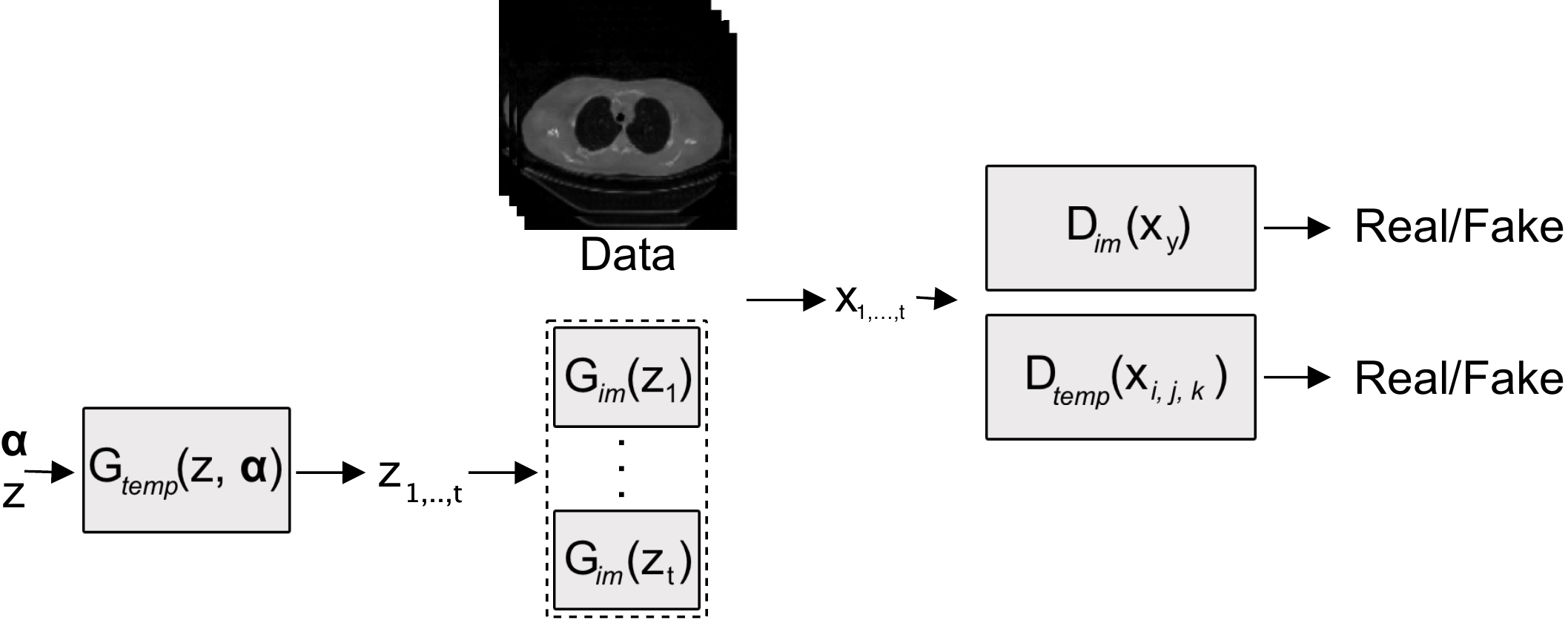}
    \caption{Schematic overview of our proposed architecture for explicit embedding in \ac{GAN} latent space. $z$ is a latent vector, $\boldsymbol\alpha$ is a set of shift magnitudes, $z_{1,...,t}$ are the shifted latent vectors, $x_{1,...,t}$ are shifted images corresponding to synthesized or real data, $x_{y}$ is the time step corresponding to the original latent vector $z$ if generated or one real image otherwise, and $x_{i,j,k}$ are three time steps, where $i,j,k\in\{1,...,t\}$.}
    \label{fig:temp_gan_base}
\end{figure}

In this section, we introduce the proposed framework. Figure \ref{fig:temp_gan_base} shows a schematic overview of our proposed model architecture.
Our base architecture takes inspiration from video \acp{GAN} and uses two discriminators. $D_{im}$ takes individual volumes, i.e., time steps, and discriminates between real and synthesized data. Thus given an underlying \ac{GAN} architecture, the discriminator can be used as $D_{im}$ without further changes. Next, the architecture has a temporal discriminator $D_{temp}$, which, given three volumes, discriminates whether they are temporally consistent. Again, this is implemented using an underlying \ac{GAN} architecture and tripling the input channels to allow for the input of temporal sequences. Further, we use two generators. $G_{im}$ is a traditional generator from a \ac{GAN} taking some latent code $z\in\mathbb R^L$, where $L$ is the latent space size, and mapping it to an image $x$ without any further changes. Finally, the temporal generator $G_{temp}$ takes a latent code $z$ and shift magnitudes $\alpha$ and shifts $z$ by magnitudes $\alpha$ along a learned linear direction $d\in \mathbb R^L$ where $L$ is the size of the latent space. These shifted latent codes are individually used as input to $G_{im}$ to generate the sequence of volumes. Thus, rather than directly generating a sequence, we embed a direction in the latent space corresponding to the desired change, and by shifting with increasing $\alpha$, we can create a set of latent codes $\Vec{z}$ corresponding to consecutive time steps of variable length. These latent codes are individually used as input to $G_{im}$ to create the desired number of time steps in data space.\\
Given the design of the proposed model, any \ac{GAN} architecture consisting of discriminator $D$ and generator $G$ can be used by adding the temporal generator $G_{temp}$ as detailed above and using the discriminator architecture twice as $D_{im}$ and $D_{temp}$ to construct an explicitly embedding \ac{GAN}.\\
We suggest the following implementation details: Based on the work of Voynov and Babenko \cite{Voynov20}, we use a direction $d$ of unit length and $\alpha \in \mathcal U[-6,6]$. As the image discriminator follows standard \ac{GAN} training, we suggest using the same loss for the image discriminator and generator that the base architecture uses.\\
For temporal consistency, we optimize using adversarial learning. Let $p_{true}$ be the distribution of real data correctly ordered w.r.t. transformation magnitude and $p_{false}$ be incorrectly ordered, and $p_z$ the latent distribution. Further, let $\boldsymbol\alpha = (\alpha_1, \alpha_2, \alpha_3)$ be any $\alpha_i\in \mathbb R$ for which $\alpha_1\le \alpha_2 \le\alpha_3$, then we define the adversarial loss objective for the temporal discriminator using the hinge loss as:
\begin{equation}
 \label{eq:temp_d_adv}
    \begin{split}
\underset{D_{temp}}{\min} \: \mathcal L_{D_{temp}} &= \underset{x\sim p_{true}}{\mathbb E} [\min(0,1-D_{temp}(x))]\\
    &+ \underset{x\sim p_{false}}{\mathbb E}[\min(0,1+D_{temp}(x))]\\
    &+ \underset{z\sim p_z}{\mathbb E}[\min(0,1+D_{temp}(G_{im}(G_{temp}(z, \boldsymbol\alpha))))]
\end{split}
\end{equation}
Given that we want to force the embedding in the latent space, we add a loss term for both $G_{im}$ and $G_{temp}$ so that $G_{temp}$ learns the direction and $G_{im}$ learns the latent representation corresponding to the desired transformation. Thus, we get:
\begin{equation}
    \label{eq:temp_g_adv}
    \underset{G_{im}, G_{temp}}{\min}\: \mathcal{L}_G =
    \underset{z,z'\sim p_z}{\mathbb E}\big [
    -D_{temp}(G_{im}(G_{temp}(z, \boldsymbol\alpha)))
    + L_{GAN} (D_{im}(G_{im}(z')))
    \big ]
\end{equation}
where $L_{GAN}$ is the applicable \ac{GAN} loss of the base architecture, and $\boldsymbol \alpha$ and $p_z$ are defined as above.\\
Intuitively, the temporal discriminator learns to discriminate based on the transformation we aim to embed. Therefore, the generators trying to fool the temporal discriminator must generate data that exhibits the correct transformation and does not change the scene (e.g., patient) markedly.\\
We evaluate image quality using visual inspection and slice-wise \ac{FID} and visual inspection for temporal consistency. The shifted images should not be of worse image quality than images resulting from directly sampled latent codes. Thus, the basic image quality measures are also used to assess the shifted images.

\section{Experiments}
\label{sec:experiments}
\subsubsection{Datasets} 
\begin{figure}[h!]
    \centering
    \includegraphics[width=\textwidth]{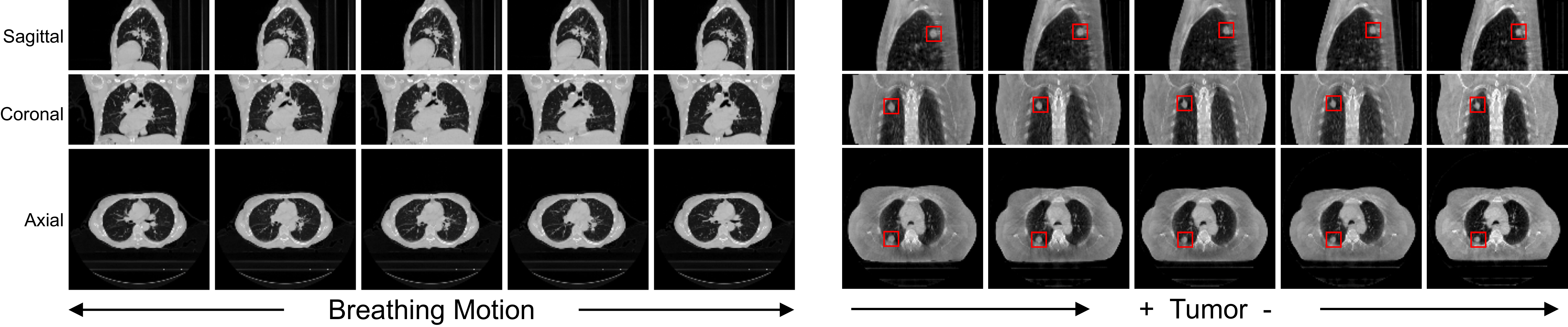}
    \caption{Examples of breathing motion and tumor regression datasets. The presented examples are after preprocessing. For breathing motion, the center volume shows the most exhaled while the ones to the left correspond to exhaling and those to the right to inhaling. For tumor regression, we estimate the slice corresponding to the center of the tumor manually. The tumor is marked with the red bounding box.}
    \label{fig:real_data}
\end{figure}

We evaluate the proposed architecture on three volumetric thoracic \ac{CT} datasets.
\begin{itemize}
    \item \ac{LIDC} \cite{Armato11}. We preprocess \ac{LIDC} by limiting the intensity range to $[-1000,2000]$ \ac{HU} and normalize to a range of $[-1,1]$ using min-max scaling. We resize the data to $128\times128\times128$ voxels to limit computational demands. To test the approach, we introduce a simple image transformation corresponding to shifts along the $x$-axis with a shift magnitude of $[-32, 32]$ voxels randomly selected.
    \item Breathing Motion. The dataset, collected at Rigshospitalet, Copenhagen, Denmark, consists of longitudinal thoracic \ac{CT} scans showing the breathing phases of $499$ non-small cell lung cancer patients treated with radiotherapy between $2009$ to $2017$. Each data point generally consists of $10$ scans, where the first five correspond to exhaling and the following five to inhaling. We only consider the scans corresponding to exhaling. We limit the range to $[-1000,500]$ \ac{HU} and normalize to a range of $[-1,1]$ using min-max scaling. We resize the data to $128\times128\times64$ voxels to limit computational demands.
    \item Tumor Regression. The final dataset consists of $256$ patients, a subset of the patients from the Breathing Motion dataset, for which at least $10$ daily treatment thoracic cone beam \ac{CT} scans were available. It shows tumor regression during radiotherapy. We apply the same preprocessing as with the breathing motion dataset.
\end{itemize}
For all three datasets, we use $90\%$ of the patients for training and $10\%$ for testing. Figure \ref{fig:real_data} shows an example of the breathing motion and tumor regression dataset.

\subsubsection{Implementation Details} We run all experiments on two Nvidia RTX A6000. We use Python $3.9$ and PyTorch $1.11.0$ for the implementation. The first author performed all visual inspections without formal education in medical image evaluation. We adapt \ac{SA-GAN} \cite{Zhang19} to volumetric data and use it as our base \ac{GAN} architecture following the parameters (e.g., learning rate and optimizer) suggested by the authors unless otherwise specified. Throughout all experiments, we sample $\alpha\sim \mathcal U[-6,6]$ with a linear direction $d$ of unit length based on Voynov and Babenko \cite{Voynov20}, and we arbitrarily choose to train for $5000$ iterations. We use a batch size of $8$ for the \ac{LIDC} dataset and a batch size of $16$ for the other two to fit the memory of the used GPUs. For the \ac{LIDC} dataset, we use a latent space size of $L=512$ and reduce it to $L=256$ for the other two, as the resolution of the datasets is halved as well.

\section{Results}
\label{sec:results}
\begin{figure}[h!]
    \centering
    \includegraphics[width=\textwidth]{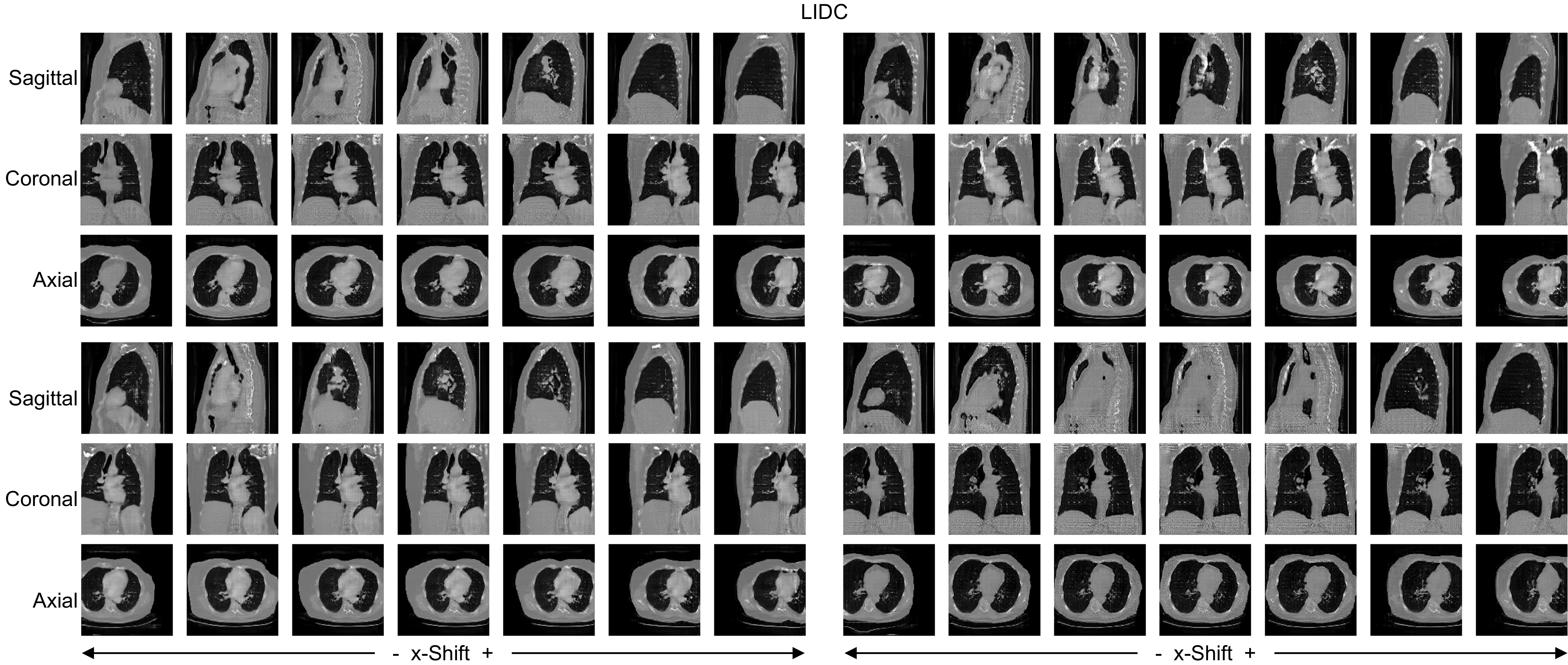}
    \caption{Four examples of generated volumes with embedded shift for the proposed model on the \ac{LIDC} dataset. The center volume corresponds to the original latent vector. All images show the center slice for the sagittal, coronal, and axial view.}
    \label{fig:temp_lidc}
\end{figure}

\begin{figure}[h!]
    \centering
    \includegraphics[width=\textwidth]{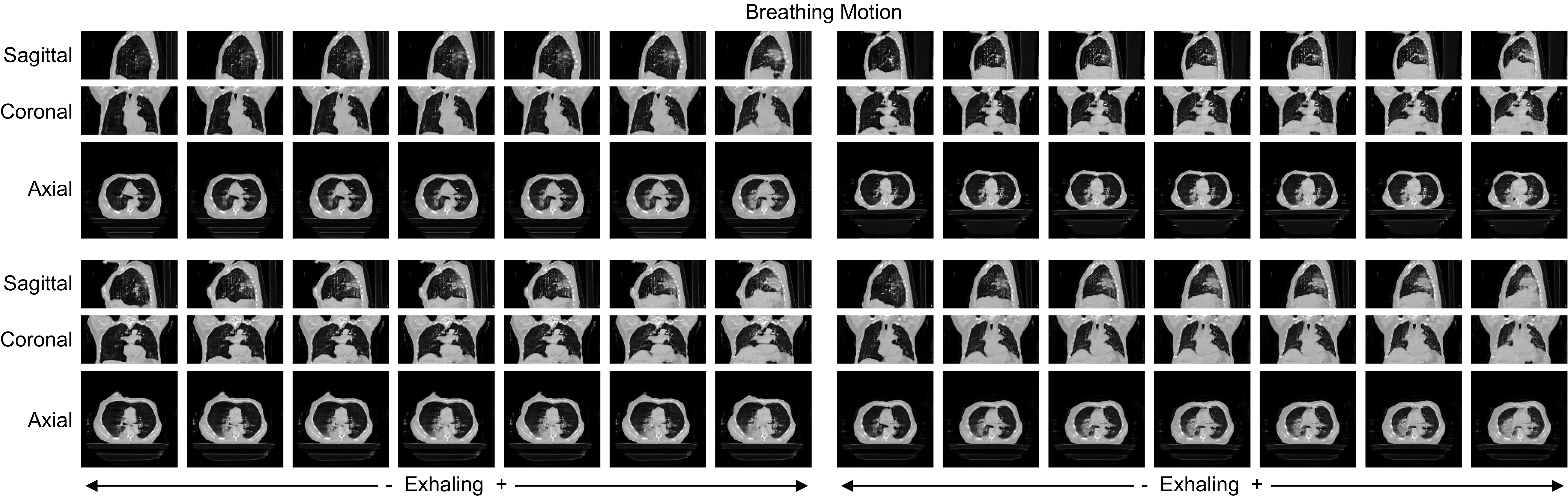}
    \caption{Four examples of generated volumes with embedded shift for the proposed model on the breathing motion dataset. The center volume corresponds to the original latent vector. All images show the center slice for the sagittal, coronal, and axial view.}
    \label{fig:temp_breath}
\end{figure}

\begin{figure}[h!]
    \centering
    \includegraphics[width=\textwidth]{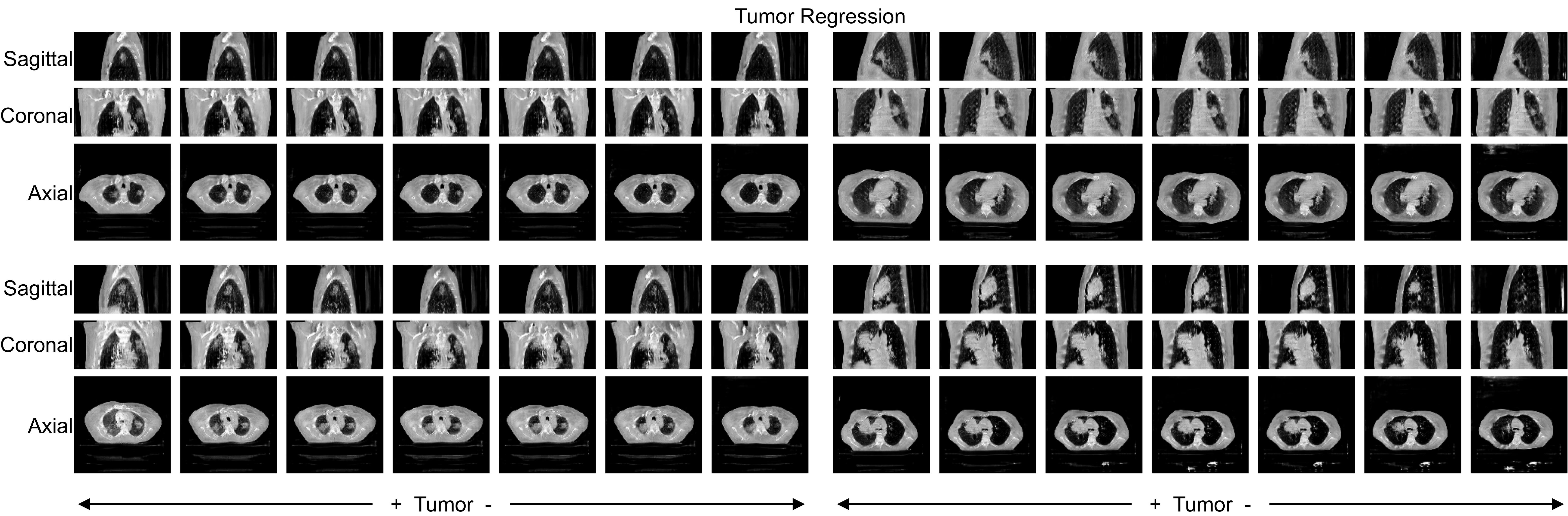}
    \caption{Four examples of generated volumes with embedded shift for the proposed model on the tumor regression dataset. The center volume corresponds to the original latent vector. For all images, we try to locate the center of the tumor for the sagittal, coronal, and axial view.}
    \label{fig:temp_tumor}
\end{figure}

We present the final, unbiased estimate of the image quality of the proposed model on all three datasets in Table \ref{tab:temp_all_im}. Note that some image transformations are more readily visible in video format. Therefore, consider the generated volumes as videos in the provided GitHub repository.
\begin{table}[h!]
    \centering
   
    \begin{tabular}{c||c|c|c}
       & \ac{FID} ax. & \ac{FID} cor. & \ac{FID} sag.\\\hline\hline
      \ac{LIDC} & $93.8\pm1.0$ & $54.3\pm 0.5$ & $30.0\pm 1.2$\\
      Breathing Motion & $139.2\pm2.7$ & $79.8\pm 1.0$ & $99.6\pm 1.4$\\
      Tumor Regression & $82.2\pm2.7$ & $42.3\pm 1.4$ & $53.4\pm 0.8$
    \end{tabular}
    
    \caption{Image Quality of the temporal \ac{GAN} trained on all three datasets. All models are trained on $90\%$ of the scans split patient-wise and evaluated on $10\%$. The \ac{FID} scores are calculated using random time steps of real and synthesized data.}
    \label{tab:temp_all_im}
\end{table}
Examples of generated volumes of the model trained on the full \ac{LIDC} data set are given in Figure \ref{fig:temp_lidc}.
The resulting volumes are of high quality, and we observe the desired image transformation embedded as a shift in the latent space. The transition between shifted images is smooth with minimal entanglement.\\
Next, we consider the breathing motion dataset. Figure \ref{fig:temp_breath} presents some generated volumes for the breathing motion dataset.
We observe good image quality with sufficient detail and realistic anatomy. Considering the embedding, we observe that breathing motion is captured well, and the temporal dependencies of breathing are embedded in the latent space. The clearest change when moving along the embedded direction is the diaphragm moving upward while exhaling. This is also the most obvious change observable in the real data. Additionally, we observe the stomach or rib cage contracting while exhaling. Lastly, we observe very high scene consistency. I.e., the generated scan of the patient does not change markedly while moving the latent code along the embedded direction. Thus, the patient's anatomy is preserved, and the changes induced by moving along the embedded direction are restricted to breathing-related changes.\\
Finally, we consider the model trained on the tumor regression dataset. We present examples of generated volumes in Figure \ref{fig:temp_tumor}.
The image quality of the generated volumes is good, showing details in the vessel, tissue, and bone structure. We observe volumes both containing and not containing tumors. Those generated volumes containing tumors have them in varied places, shapes, and sizes. If tumors are present in the generated volumes, the temporal generation results in tumors shrinking in size. I.e., the model successfully embeds temporal tumor regression in image space as a linear direction in the latent space. When traversing the embedded direction, there is minimal change to the volumes other than the reduction in tumor size. Further, no clear change exists if no tumor is present in the volume. Thus, the direction models tumor regression in a disentangled manner.

\section{Discussion}
\label{sec:discussion}
\subsubsection{Temporal Generation}
We can use a simple learned direction to generate temporal sequences of data using only a single non-temporal generator, which, to the best of our knowledge, has not been shown previously. Our proposed method of jointly training such a direction and the \ac{GAN} shows distinct benefits over discovering directions in pretrained generators.
From visible inspection, our method shows almost no entanglement and ensures that even complex transformations, such as tumor regression, are enforced to be present as a linear direction in the latent space.\\
Our model shows high-quality synthetic data with controllable enforced image transformations with smooth continuous generation on three datasets. Additionally, in particular, on the breathing motion and tumor regression dataset, we observe clearer changes than observed in the data (e.g., movement of the diaphragm in the breathing motion dataset). This indicates that the proposed method isolates the signal corresponding to temporal development well. Compared to other temporal generation approaches, our model does not require conditioning of the generator or a sequential generator. As a consequence, we can easily vary the architecture and benefit from advances in \ac{GAN} architectures that are to come.\\
The results we observe on the tumor regression dataset deserve the most attention. While traditional tumor regression models might be more patient- and therapy-specific \cite{Huang10}, the scale we operate on is novel. Furthermore, unlike previous methods, we generate entire volumes and show that we can model tumor regression as part of the image generation process.

\subsubsection{Limitations}
We use the parameters used for \ac{SA-GAN}. While this is likely a good choice for the image generation component of our proposed model, the temporal aspects could perform better with different parameters and architecture choices. Moreover, as there is little prior investigation into evaluating temporal \acp{GAN}, we needed to devise our evaluation strategy, further development in this area would likely be beneficial.

\subsubsection{Impact \& Future Work}
\begin{figure}[h!]
    \centering \includegraphics[width=\textwidth]{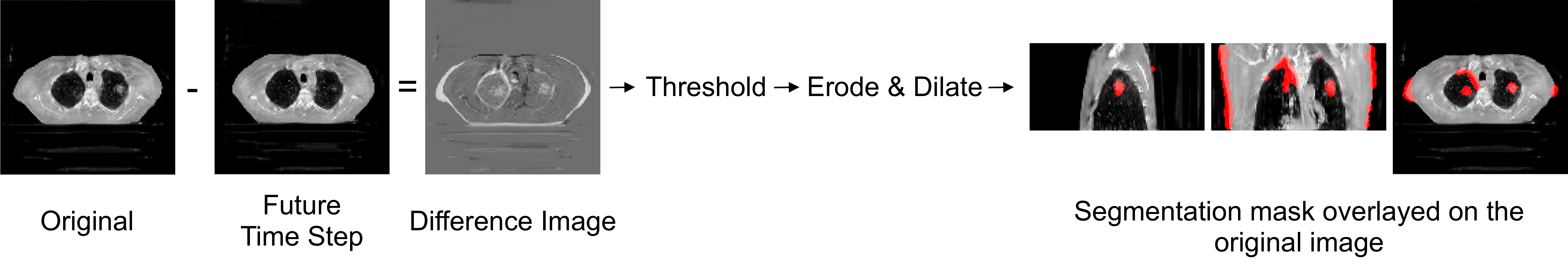}
    \caption{Proof of concept of unsupervised tumor segmentation using our method. We generate a future time point to a given volume, take the difference image, and threshold it with a threshold of $0.2$. Then we apply two erosion followed by two dilation operations to remove noise and are left with the segmentation mask. Next to the direct application to tumor segmentation, the difference image might also visualize treatment effects, such as weight loss, which is a common side effect.}
    \label{fig:seg_example}
\end{figure}

As our method is trained based only on the order of the transformation magnitude, it is reasonable to assume that it can be applied to any transformation where such an ordering exists. Further, our method provides many practical applications to downstream machine learning tasks by providing a method to synthesize controllable data, e.g., with or without tumor, in a domain where annotations are costly and difficult to obtain. We provide a proof of concept of unsupervised tumor segmentation using our method in Figure \ref{fig:seg_example}.
\\
Further practical applications directly benefiting medical image analysis will likely arise with further investigation of our method. Given the clear and isolated signal we observe, natural applications of our work could be in the visualization and understanding of changes happening in temporal sequences. Additionally, investigating how much the sampled temporal development reflects patient-specific as opposed to therapy-specific aspects would offer valuable insights. Finally, we see the future investigation of embedding non-temporal dependencies as one of the most promising potential applications. Embedding transformations across patients, e.g., disease severity, would allow for further fine-grained control over data synthesis in the medical domain.

\section{Conclusion}
\label{sec:conclusion}
In this work,  we investigate the possibility of explicitly embedding temporal dependencies in the latent space of generative models to generate longitudinal volumetric data. We generate controllable longitudinal data with minimal entanglement. Further, linear directions in the latent space are sufficient to generate temporal sequences from a non-temporal generator. Due to the simplicity of the linear latent walk, we can generate continuous and smooth sequences of varying lengths, unlike other suggested temporal \acp{GAN}.\\
We show that our framework can generate complex temporal dependencies, e.g., breathing motion or tumor regression, as part of the image synthesis task on medical data. The method could potentially improve unsupervised tumor segmentation, disease-aware image augmentation, and radiotherapy planning. Further, our method can embed any temporal dependency with limited supervision and thus provides further usefulness beyond what we explore in this work.

\subsubsection{Acknowledgements} The authors acknowledge the National Cancer Institute and the Foundation for the National Institutes of Health, and their critical role in the creation of the free publicly available LIDC/IDRI Database used in this study. The authors would like to thank Anna Kirchner for help in preparation of the manuscript. Jens Petersen is partly funded by research grants from the Danish Cancer Society (grant no. R231-A13976) and Varian Medical Systems.


\bibliography{ref}
\bibliographystyle{splncs04}

\end{document}